\documentclass[journal]{IEEEtai}

\usepackage[colorlinks,urlcolor=blue,linkcolor=blue,citecolor=blue]{hyperref}

\usepackage{color,array}
\usepackage{hyperref}
\usepackage{graphicx} 
\usepackage{amsmath}
\usepackage{booktabs}
\usepackage{multirow}
\usepackage{subcaption} 
\usepackage{booktabs}  
\usepackage{multirow}
\usepackage{amsfonts}
\usepackage{float}

\begin{document}

\title{Detecting the Undetectable: Combining Kolmogorov-Arnold Networks and MLP for AI-Generated Image Detection  } 
\author{
    \IEEEauthorblockN{Taharim Rahman Anon}
    \IEEEauthorblockA{
        \textit{Barisal Information Technology College (BITC)}\\
        Barisal, Bangladesh\\
        tahrim.anon21@gmail.com }\\
    \and
    \IEEEauthorblockN{Jakaria Islam Emon}
    \IEEEauthorblockA{
        \textit{American International University Bangladesh (AIUB) }\\
        Dhaka, Bangladesh\\
        21-92037-2@student.aiub.edu}
}
\maketitle
\begin{abstract}
As  artificial intelligence progresses, the task of distinguishing between real and AI-generated images is increasingly complicated by sophisticated generative models. This paper presents a novel detection framework adept at robustly identifying images produced by cutting-edge generative AI models, such as DALL-E 3, MidJourney, and Stable Diffusion 3. We introduce a comprehensive dataset, tailored to include images from these advanced generators, which serves as the foundation for extensive evaluation. we propose a classification system that integrates semantic image embeddings with a traditional Multilayer Perceptron (MLP). This baseline system is designed to effectively differentiate between real and AI-generated images under various challenging conditions. Enhancing this approach, we introduce a hybrid architecture that combines Kolmogorov-Arnold Networks (KAN) with the MLP. This hybrid model leverages the adaptive, high-resolution feature transformation capabilities of KAN, enabling our system to capture and analyze complex patterns in AI-generated images that are typically overlooked by conventional models.  In out-of-distribution testing, our proposed model consistently outperformed the standard MLP across three out of distribution test datasets, demonstrating superior performance and robustness in classifying real images from AI-generated images with impressive F1 scores.  
\end{abstract}
\section{Introduction}

\IEEEPARstart{T}{he} he rapid advancement in artificial intelligence (AI) has led to significant progress in image generation technologies, resulting in highly realistic synthetic images \cite{pmlr-v139-ramesh21a}. While advancements in image generation \cite{NEURIPS2021_49ad23d1} technologies bring numerous benefits, they also present significant challenges in misinformation and digital forensics \cite{Naitali2023}.  Maintaining the integrity of visual media relies crucially on the ability to differentiate between AI-generated and real images. Lu et al. \cite{lu2024seeing} underscore this challenge, revealing in their study significant difficulties even among human observers. Utilizing a large-scale fake image dataset named Fake2M and conducting a human perception evaluation called HPBench, they found a notable misclassification rate of 38.7\%, indicating that humans struggle considerably to distinguish real photos from AI-generated ones.

Several studies have focused on developing methods to detect AI-generated images, each with its own strengths and limitations. Epstein et al. \cite{epstein_online_2023} evaluate classifiers trained incrementally on 14 AI generative models arranged by release date, to discern real from AI-generated images. The study highlights the classifiers' generalization capabilities to newer AI models but notes a limitation: advance generative models like DALL-E and MidJourney, especially those without public APIs, are not included. This exclusion leads to observed performance drops with major architectural changes in emerging models, emphasizing the need for continual updates and retraining of these classifiers .

Chen et al. present a novel detection method using noise patterns from a single simple patch (SSP) to distinguish between real and AI-generated images, capitalizing on the noise discrepancies that AI-generated images often overlook \cite{chen_single_2023}. Their methodology includes the SSP Network, which extracts and analyzes noise patterns using SRM filters and a ResNet50 classifier. Despite these advancements, the study notes significant challenges with very low-quality images, where performance deteriorates, particularly with images having compression quality below 90. However, a notable limitation, as observed, is the exclusion of advanced generative models in the datasets used, which might impact the method’s applicability to newer or more sophisticated AI image generators. 

Wang et al. proposed DIRE for detecting diffusion-generated images, demonstrating strong generalization and robustness, but noted the computational cost of their inversion and reconstruction process \cite{wang_dire_2023}.  Yan et al. developed the Chameleon dataset to challenge existing detectors and proposed the AIDE model, which showed competitive performance but highlighted ongoing detection challenges and higher computational costs \cite{yan_sanity_2024}. Martin-Rodriguez et al. used  Photo Response Non-Uniformity (PRNU) and Error Level Analysis for pixel-wise feature extraction combined with CNNs, achieving high accuracy but noted the limitations with JPEG format dependency and potential manipulation of PRNU \cite{martin-rodriguez_detection_2023}. Chai et al. focused on local artifacts for detection, showing strong generalization but acknowledged preprocessing artifacts and adversarial vulnerability \cite{chai_what_2020}. Park et al. provided a comprehensive comparison and visualization of various detection methods, demonstrating the strengths and weaknesses of each, particularly their robustness to augmentations like JPEG compression and Gaussian blur \cite{park_performance_2024}.

Building upon the existing research landscape, this study introduces a robust approach to tackle the challenge of generator variance—effectively distinguishing between images generated by diverse AI technologies. Our method not only addresses common limitations found in previous works, such as the inability to handle newer or more sophisticated generative models, but also demonstrates enhanced adaptability across a wide range of image qualities and generator architectures. The key contributions of our work are detailed as follows:
\begin{itemize}
    \item We have compiled a  dataset that includes images generated by the latest models, such as DALL-E 3 \cite{betker2023improving}, MidJourney 6, and Stable Diffusion 3 Ultra \cite{esser2024scaling}. This dataset is crucial for developing a model that is well-versed in the nuances of the most recent AI image generators, ensuring robust detection capabilities across a spectrum of generative technologies.
    \item We proposed a classification system as a baseline approach that integrates semantic image embeddings with a MLP. This combination is specially designed to effectively differentiate between real and AI-generated images, even under challenging conditions such as low image quality or subtle image features. 
    \item We proposed a hybride architecture that seamlessly integrates Kolmogorov-Arnold Networks (KAN) with a traditional MLP to further enhance the classification efficacy. This approach leverages the unique capabilities of KAN for adaptive, high-resolution feature transformation, allowing our system to capture and analyze intricate patterns and complexities in AI-generated images that are often missed by conventional models. 
    \item In out-of-distribution testing, proposed model consistently outperformed the standard MLP model across three test datasets: Real vs. DALL-E 3, Real vs. Midjourney 5, and Real vs. Adobe Firefly. The F1-Scores for the Hybrid KAN MLP were 0.94, 0.94, and 0.91, respectively, demonstrating superior performance and robustness in discriminating real from AI-generated images. 
\end{itemize}

\section{Data preparation}
To develop our AI-generated image detection framework, we utilize a subset of the RAISE dataset \cite{DangNguyen2015RAISE}. Designed specifically for digital forgery detection algorithms, the RAISE dataset consists of 8,156 high-resolution, uncompressed RAW images that are guaranteed to be camera-native. This characteristic ensures no prior processing has altered the images, making it ideal for authenticity testing. 

\subsection{Data Generation for Training}
\begin{figure}[h]
    \centering
    \includegraphics[width=\columnwidth]{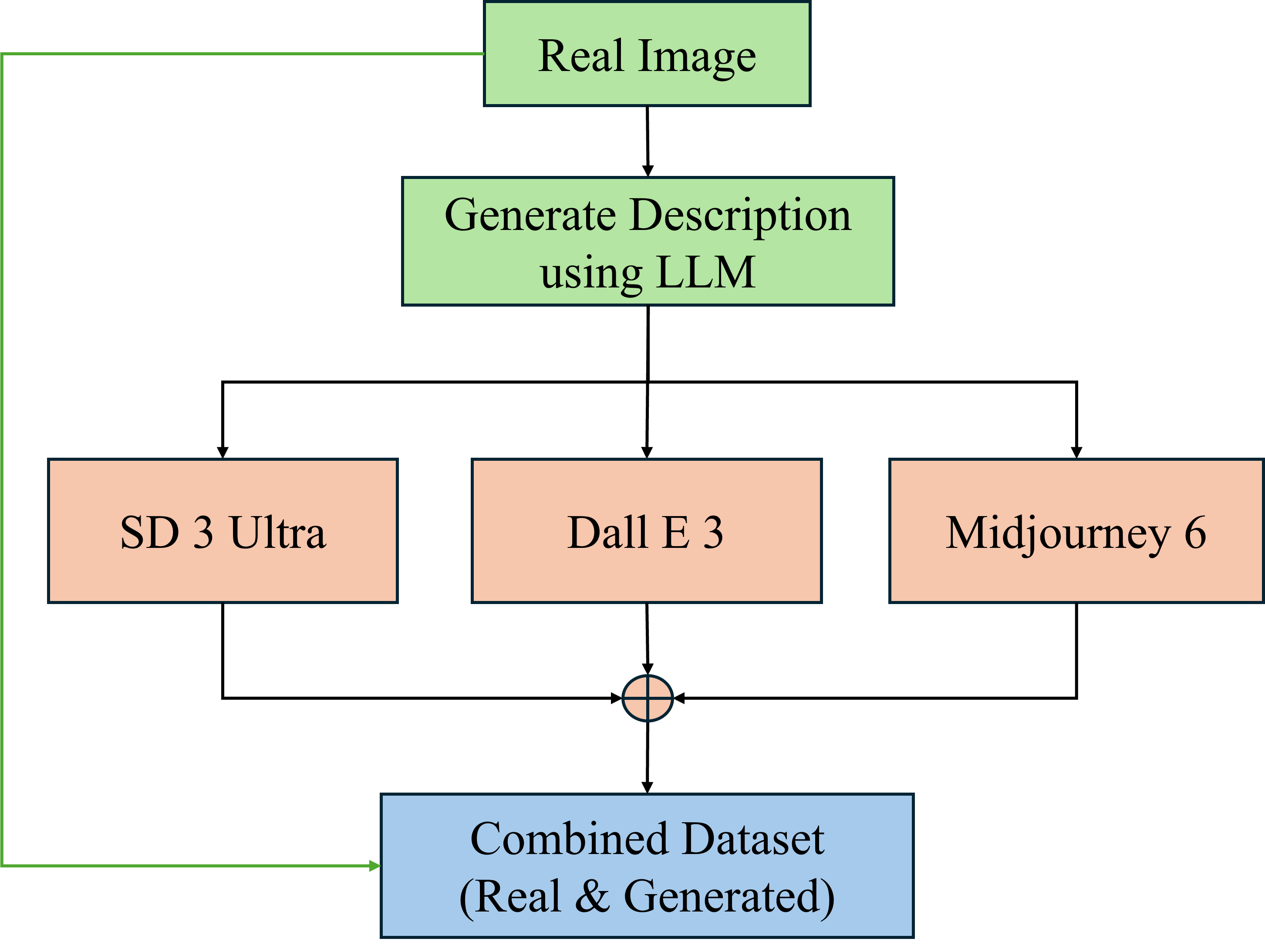}
    \caption{Illustrates of  data generation process in our study. Starting with real images, descriptions are generated using a language model. These descriptions are then used to generate AI images using state-of-the-art models such as Stable Diffusion 3 Ultra (SD 3 Ultra), DALL-E 3, and MidJourney 6. }
    \label{fig:ds_gen}
\end{figure}

\begin{figure*}[htbp]
    \centering
    \includegraphics[width=1\linewidth]{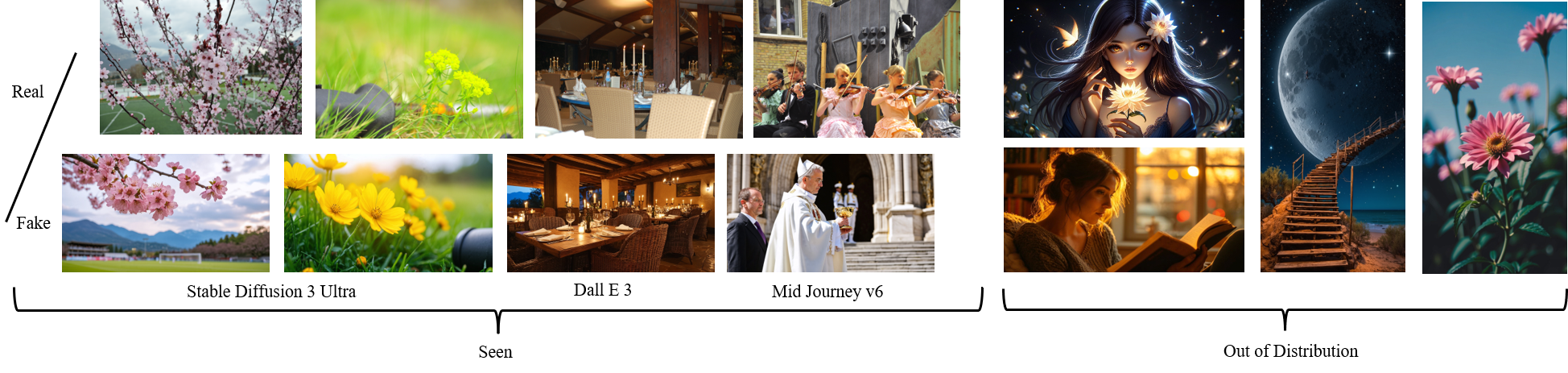}
    \caption{Generated images from given real image descriptions using generative AI models.}
    \label{fig:ai_images}
\end{figure*}

To create a robust model for distinguishing real from AI-generated images, we implemented an extensive data generation process utilizing paid image generation services to address the research gap in generated image detection tasks. Our approach involved the following steps as outlined in figure \ref{fig:ds_gen}. For this study, we selected 1000 images from the RAISE dataset, referred to as RAISE 1k. This subset provides a diverse range of real-world imagery, enabling robust testing of our detection framework under varied conditions.  For each real image, we generated a detailed description using a large language model (OpenAI's GPT-4.) These descriptions encapsulate the key visual elements and context of the images. The generated descriptions were split into three folds and used to create synthetic images through three advanced image generation models: Stable Diffusion 3 Ultra (SD 3 Ultra), DALL-E 3, and MidJourney 6. Each model adds its unique characteristics to the generated images, enhancing the diversity of the dataset as shown in figure \ref{fig:ai_images}. After that real images and the AI-generated images were combined to form a comprehensive dataset.  This process ensures our model is exposed to a balanced mix of real and synthetic images during training, as summarized in Table \ref{tab:image_summary}. 

\begin{table}[h]
    \centering
    \caption{Summary of Real and AI-Generated Images}
    \label{tab:image_summary}
    \begin{tabular}{lc}  
        \toprule
        \textbf{Category} & \textbf{Number of Images} \\
        \midrule
        Real Images (RAISE 1k) & 1000 \\
        AI-Generated Images (SD 3 Ultra) & 340 \\
        AI-Generated Images (DALL-E 3) & 333 \\
        AI-Generated Images (MidJourney 6) & 333 \\
        \midrule
        \textbf{Total Combined Dataset} & \textbf{2000} \\
        \bottomrule
    \end{tabular}
\end{table}

\begin{table}[h]
\centering
\caption{Captions generated for images in the dataset using large language model.}
\label{tab:captions}
\begin{tabular}{@{}lp{6cm}@{}}  
    \toprule
    \textbf{Filename} & \textbf{Caption} \\
    \midrule
    r000da54ft.png & Branches of a cherry blossom tree in bloom, with a sports field and mountains in a blurred background. \\
    r001d260dt.png & Low-angle view of a brick building with two windows, one with a metal grille and red flowers, the other with wooden shutters. \\
    r002fc3e2t.png & A large fountain with high-spraying water, surrounded by stone and a pond, with people and lush greenery in the background. \\
    \bottomrule
\end{tabular}
\end{table}

This thorough data generation process ensures that our detection model is trained on a diverse and representative dataset, enhancing its ability to generalize across different types of images and generative models. 

\subsection{Out of Distribution Test Data for Model Validation} 
To ensure the robustness and reliability of our model, particularly its ability to generalize beyond the training data, we have incorporated a comprehensive Out of Distribution (OOD) test dataset. This dataset is specifically designed to challenge the model with new, unseen examples that were not present during the training phase. The purpose of this OOD test data is to evaluate the model's invariant classifications—its ability to maintain consistent and accurate predictions across varying inputs that deviate from the typical distribution of the training set.  A set of 500 real images collected randomly provides a diverse basis for testing. These images ensure that the model can effectively recognize authentic visuals that have not been previously encountered during training.  To test the model’s ability to detect newly emerging generative models, we include 500 images generated by Adobe Firefly, a different generative model from those used in training. This tests the adaptability of our detection framework. Images generated by DALL-E 3 and MidJourney 5, amounting to 1000 images (500 each), serve to further diversify the test conditions. This inclusion helps ascertain the model's performance against a variety of generative techniques it has seen but under new parameters.
\begin{table}[h]
    \centering
    \caption{Summary of Out of Distribution AI-Generated Images for Test Dataset}
    \label{tab:ood_image_summary}
    \begin{tabular}{lc}  
        \toprule
        \textbf{Category} & \textbf{Number of Images} \\
        \midrule
        Real Images (Randomly Collected) & 500 \\
        AI-Generated Images (Adobe Firefly) & 500 \\
        AI-Generated Images (DALL-E 3) & 500 \\
        AI-Generated Images (MidJourney 5) & 500 \\
        \midrule
        \textbf{Total Combined Dataset} & \textbf{2000} \\
        \bottomrule
    \end{tabular}
\end{table}
This strategy enhances our model's ability to generalize and provides a stringent test of its effectiveness across a broad spectrum of real and synthetic images. The OOD test dataset plays a critical role in validating the model under varied and challenging conditions.

\section{Methodology}
Our methodology is designed within a comprehensive framework aimed at effectively distinguishing real images from AI-generated ones. This structured approach comprises four critical stages: data generation, embedding generation, classifier training, and rigorous model evaluation. Initially, we create a dataset with a combination of real images and generated images by leveraging a large language model and image generation models. This dataset forms the foundation for generating semantic image embeddings, which capture the essential visual and contextual information of the images. These embeddings serve as the input for training two types of classifiers: a standard MLP as our baseline and our proposed Hybrid KAN-MLP. We then undertake a thorough evaluation of these classifiers using  out-of-distribution data. This evaluation focuses on key metrics such as the AUC ROC, F1 score, precision, and recall to assess the classifiers' performance comprehensively. Our methodology not only highlights the potential applications of these classifiers in real-world scenarios but also ensures a robust analysis of their effectiveness in distinguishing between real and generated image. 
\begin{figure*}
    \centering
    \includegraphics[width=1\linewidth]{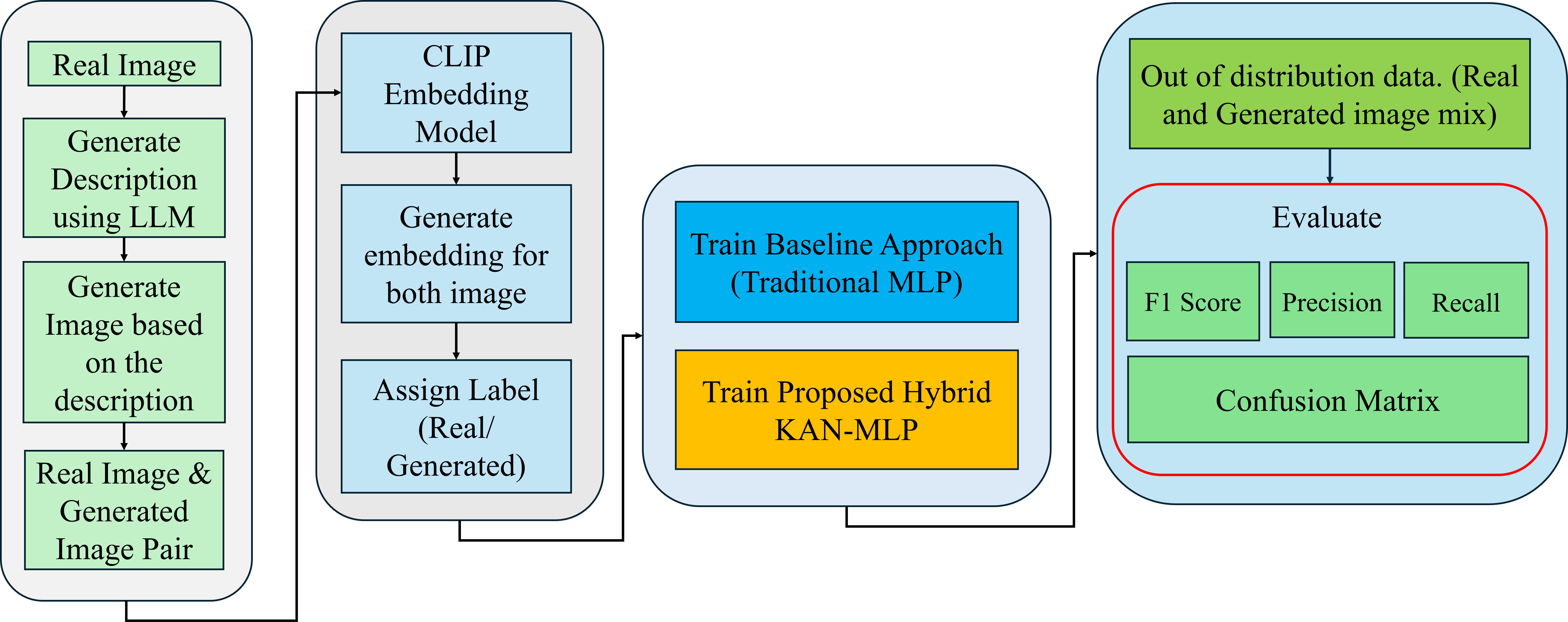}
    \caption{Proposed Methodology for AI-Generated Image Detection Framework. }
    \label{fig:enter-label}
\end{figure*}

\subsection{Semantic Image Embedding Generation}
In our study, image embeddings \cite{Tsai2011} play a pivotal role in distinguishing between real and AI-generated images. An image embedding is a compact, high-dimensional vector representation of an image, denoted as:
\begin{equation}
    \mathbf{x} \in \mathbb{R}^{512}
\end{equation}
where \(\mathbf{x}\) represents the embedding vector with a dimensionality of 512. This vector transforms the raw visual content of the image—its colors, textures, shapes, and semantic meanings—into a numerical format that deep learning algorithms can process.
The process of generating these embeddings can be mathematically represented as:
\begin{equation}
    \mathbf{x} = f_{\text{CLIP}}(\text{image})
\end{equation}
where \(f_{\text{CLIP}}\) denotes the embedding generation function implemented by the CLIP model, which maps an input image to its corresponding embedding in a structured vector space. This transformation captures the essence of the image's visual information, allowing for efficient comparison and analysis of images on a scale that is not feasible with raw pixel data.
We utilize the CLIP (Contrastive Language-Image Pretraining) model \cite{radford2021learning} to generate these embeddings for each image in our dataset, whether real or AI-generated. The CLIP model is specifically designed to produce embeddings that not only encapsulate the visual information but also align this information with textual descriptions, making the embeddings particularly rich and useful for varied AI tasks. The embeddings generated by the CLIP model serve as input features for the subsequent classification tasks.

\subsection{Designing of the Hybrid KAN-MLP Classifier }
The proposed Hybrid KAN-MLP Classifier represents a synthesis of the adaptive feature transformation of Kilimongrove Arnold Networks (KAN) and the structural integrity of a MLP. This innovative classifier is anchored by the \textbf{KANLinear} module, a cornerstone of our approach, which utilizes advanced spline-based techniques for precise and high-resolution feature mapping. This module not only enhances the model's ability to capture and analyze complex nonlinear relationships in the data but also significantly boosts its accuracy and generalizability across varied imaging contexts. The mathematical representation of the transformation function applied by the KANLinear module is given by:
\begin{equation}
    f(\mathbf{x}) = \sum_{i=1}^{N} w_i B_i(\mathbf{x}, \boldsymbol{\theta}),
\end{equation}
where \(B_i(\mathbf{x}, \boldsymbol{\theta})\) are adaptive spline basis functions, parameterized by \(\boldsymbol{\theta}\), and \(w_i\) are the learned weights.

The spline basis functions within the KANLinear module are constructed using spline interpolation, mathematically defined as follows:
\begin{equation}
    B_i(\mathbf{x}, \boldsymbol{\theta}) = \sum_{j} \theta_{ij} \cdot \text{spline\_basis}(\mathbf{x}, \text{grid}_j, \text{order}),
\end{equation}
where \(\text{spline\_basis}\) denotes the spline basis function applied at points defined by \(\text{grid}_j\) with a spline order specified by \(\text{order}\). Here, the grid and spline order are configured to 10 and 3, respectively, to balance flexibility and robustness, enabling the model to fit a variety of data patterns without overfitting. A uniform grid is established across the input feature space, ranging from -1 to 1. The transformation across this grid involves a small epsilon value to ensure smooth transitions:
\begin{equation}
    \epsilon = \min(\Delta \text{grid}), \quad \text{where } \Delta \text{grid} = \text{grid}_{i+1} - \text{grid}_i.
\end{equation}
Furthermore, the module incorporates both base and spline weights for linear and nonlinear transformations, respectively, enhancing adaptability to diverse input distributions. Initial transformations use a Sigmoid Linear Unit (SiLU) for non-linear activation:
\begin{equation}
    \text{SiLU}(x) = \frac{x}{1 + e^{-x}}.
\end{equation}
Regularization within the module is managed through a 50\% dropout rate, and batch normalization is utilized to stabilize the learning process:
\begin{equation}
    \text{BatchNorm}(x) = \gamma \left( \frac{x - \mu}{\sigma} \right) + \beta,
\end{equation}
where \(\mu\) and \(\sigma\) are the mean and standard deviation of the batch, and \(\gamma\), \(\beta\) are learnable parameters.

After the KANLinear transformation, the classifier structure includes a fully connected layer that maps transformed features to the output space, accompanied by a ReLU activation to introduce additional non-linearity. A final sigmoid activation function produces output probabilities suitable for binary classification tasks:
\begin{equation}
    \sigma(z) = \frac{1}{1 + e^{-z}}.
\end{equation} 
\subsection{MLP Classifier }
In addition to our hybrid model, we employ a standalone MLP classifier as a comparative baseline within our study. The MLP is designed with multiple dense layers, formulated to learn and classify images directly from the embeddings generated by the CLIP model. The architecture of the MLP can be mathematically represented as follows:
\begin{equation}
    \mathbf{h}_1 = \sigma(\mathbf{W}_1 \mathbf{x} + \mathbf{b}_1)
\end{equation}
\begin{equation}
    \mathbf{h}_2 = \sigma(\mathbf{W}_2 \mathbf{h}_1 + \mathbf{b}_2)
\end{equation}
\begin{equation}
    \vdots
\end{equation}
\begin{equation}
    \mathbf{y} = \text{softmax}(\mathbf{W}_n \mathbf{h}_{n-1} + \mathbf{b}_n)
\end{equation}
Where:
\begin{itemize}
    \item $\mathbf{x}$ is the input embedding vector derived from the CLIP model.
    \item $\mathbf{h}_i$ represents the hidden layer outputs.
    \item $\mathbf{W}_i$ and $\mathbf{b}_i$ are the weights and biases of the $i^{th}$ layer, respectively.
    \item $\sigma$ denotes the activation function, typically ReLU (Rectified Linear Unit) for intermediate layers.
    \item $\text{softmax}$ is the activation function used in the output layer to produce the probability distribution over the classes.
\end{itemize}
The simplicity of the MLP structure not only allows us to validate the added value of the more complex Hybrid KAN-MLP model but also provides a straightforward and effective benchmark for performance comparison. This approach allows us to highlight the capability of the hybrid model in handling more intricate image data complexities, underlining the enhancements brought about by the integration of KAN techniques within the neural network architecture.

\subsection{Training}
To train the both classifier, we utilize binary cross-entropy as the loss function. The Adam optimizer is chosen for its robustness in managing sparse gradients and optimizing the training process, particularly when dealing with the non-linear transformations and high-dimensional feature space. 

\subsection{Evaluation Metrics}

To assess the performance of our trained model, we employ a variety of metrics that provide insights into different aspects of model accuracy and effectiveness. These metrics are crucial for understanding how well the model discriminates between real and AI-generated images, ensuring robustness and reliability in practical applications. Below, we detail the specific metrics used to evaluate our model: 
\begin{itemize}
    \item \textbf{Confusion Matrix}: The confusion matrix is used to describe the performance of the classification model on a set of test data for which the true values are known. It includes the values for true positives (\(TP\)), true negatives (\(TN\)), false positives (\(FP\)), and false negatives (\(FN\)), giving a comprehensive understanding of the model's performance across different classes. The matrix can be represented as:
    \begin{equation}
    \begin{pmatrix}
    TP & FP \\
    FN & TN
    \end{pmatrix}
    \end{equation}
    where the rows represent the actual classes and the columns represent the predicted classes. 
    
    \item \textbf{Precision and Recall}: Precision measures the proportion of true positive predictions among all positive predictions made by the model, while recall (sensitivity) measures the proportion of true positives identified out of all actual positives. These metrics are calculated as follows:
    \begin{equation}
    \text{Precision} = \frac{TP}{TP + FP}
    \end{equation}
    \begin{equation}
    \text{Recall} = \frac{TP}{TP + FN}
    \end{equation}
    
    \item \textbf{F1-Score}: The F1-Score is the harmonic mean of precision and recall, providing a balance between the two metrics. It is calculated as:
    \begin{equation}
    \text{F1-Score} = 2 \times \frac{\text{Precision} \times \text{Recall}}{\text{Precision} + \text{Recall}}
    \end{equation}
    
    \item \textbf{ROC Curve and AUC}: 
    The Receiver Operating Characteristic (ROC) curve is a graphical plot used to illustrate the diagnostic ability of a binary classifier as its discrimination threshold is varied. The curve is created by plotting the True Positive Rate (TPR) against the False Positive Rate (FPR) at various threshold settings. The TPR and FPR are defined as:
    \begin{equation}
    TPR = \frac{TP}{TP + FN}
    \end{equation}
    \begin{equation}
    FPR = \frac{FP}{FP + TN}
    \end{equation}
    The Area Under the Curve (AUC) of the ROC represents the measure of the ability of the classifier to discriminate between the classes and is used as a summary of the ROC curve. The higher the AUC, the better the model is at predicting 0 classes as 0 and 1 classes as 1. An AUC of 0.5 suggests no discrimination ability (random guessing), while an AUC of 1.0 indicates perfect discrimination.
\end{itemize}

\section{Results and Discussion}

\begin{figure*}[t]
    \centering
    \begin{subfigure}[b]{0.32\textwidth}
        \centering
        \includegraphics[width=1\linewidth]{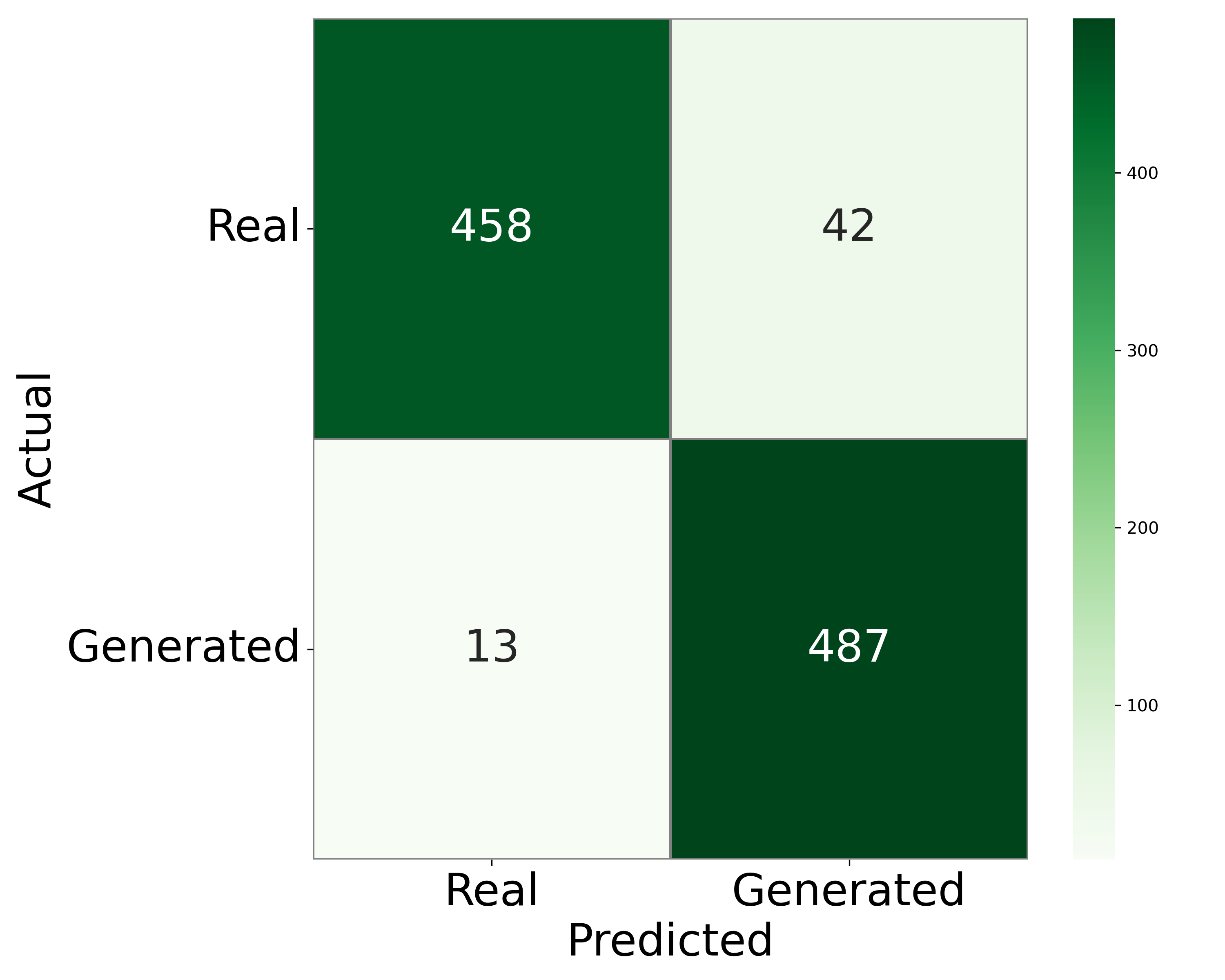}
        \caption{}
    \end{subfigure}%
    \begin{subfigure}[b]{0.32\textwidth}
        \centering
        \includegraphics[width=1\linewidth]{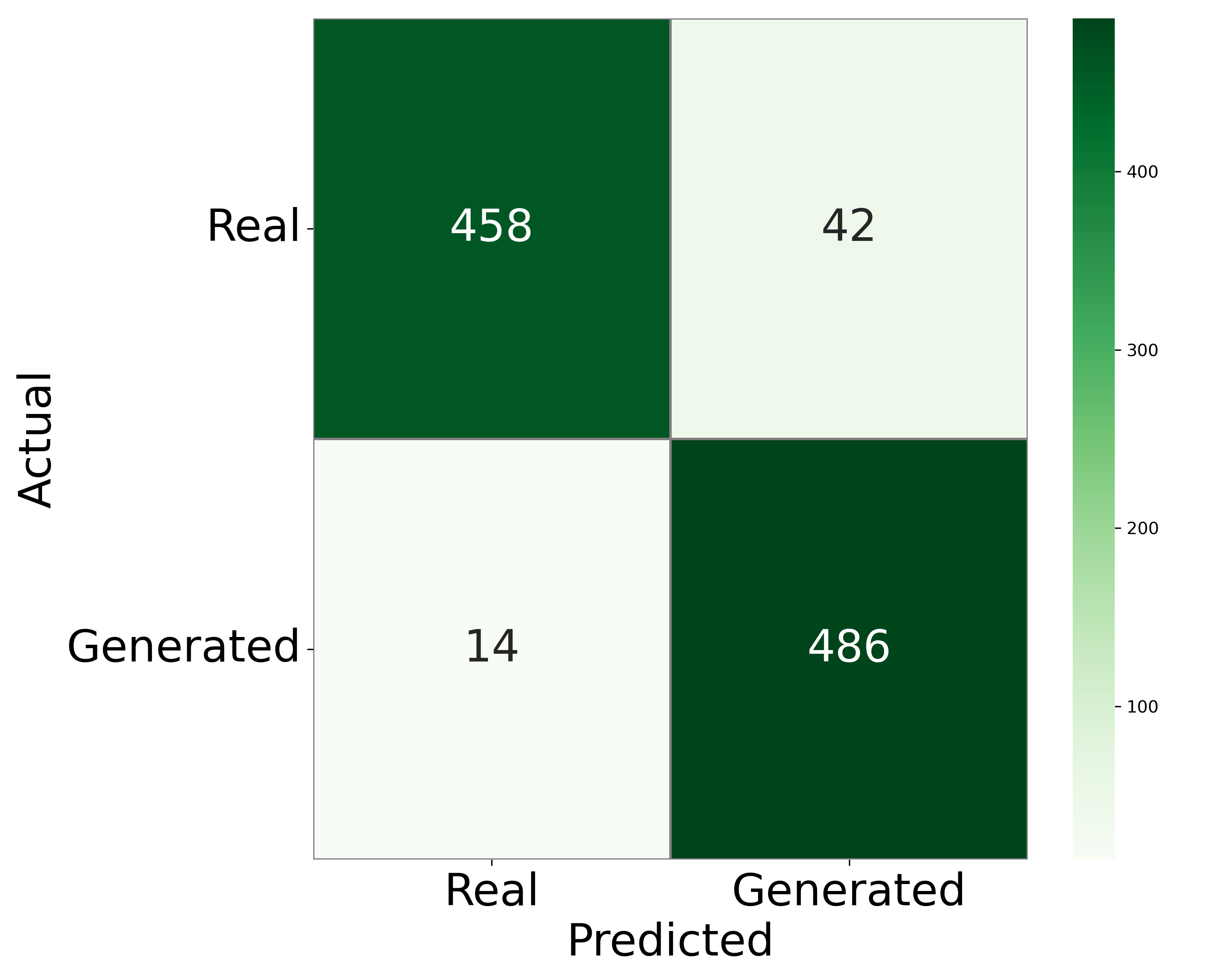}
        \caption{}
    \end{subfigure}%
    \begin{subfigure}[b]{0.32\textwidth}
        \centering
        \includegraphics[width=1\linewidth]{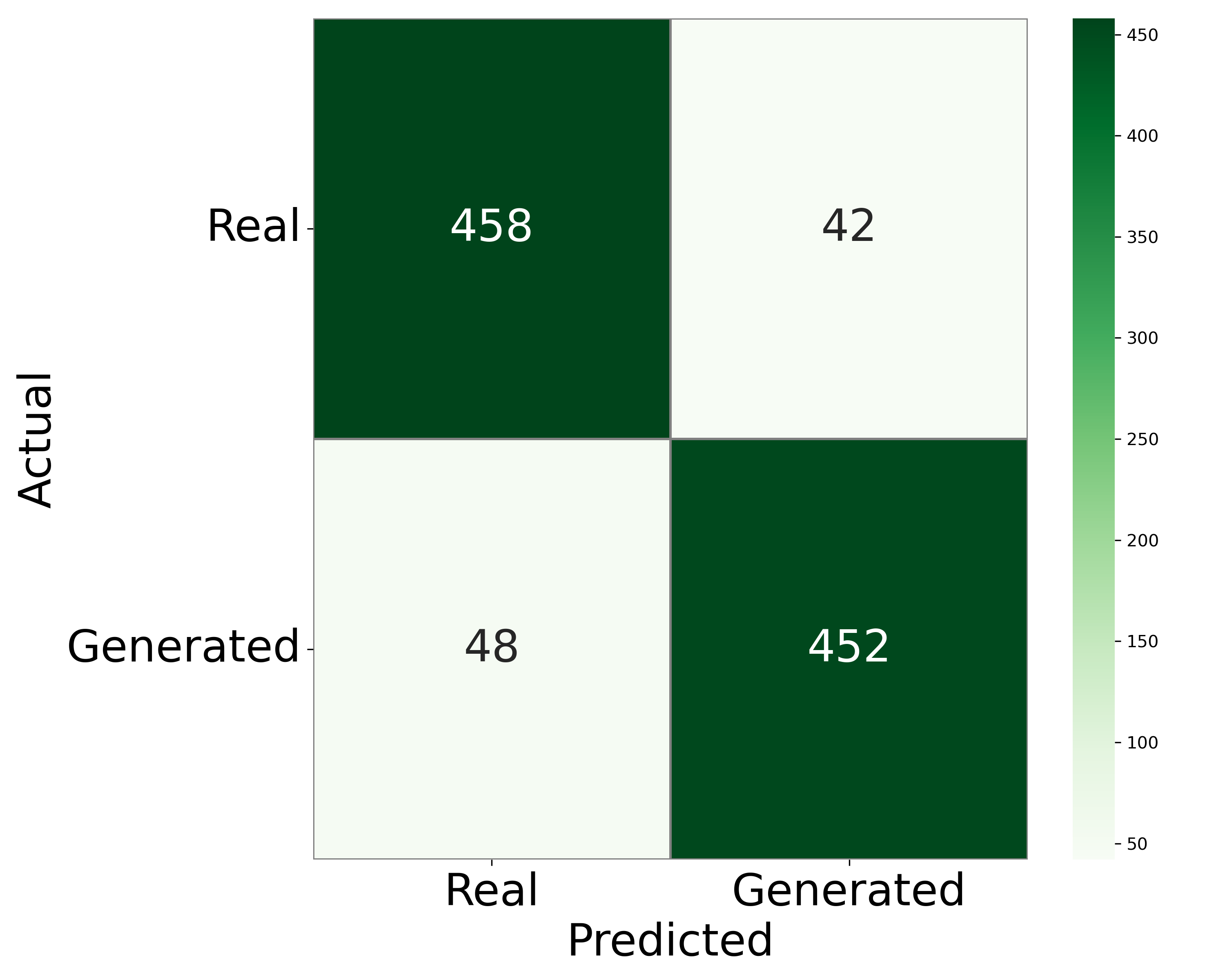}
        \caption{}
    \end{subfigure}
    \begin{subfigure}[b]{0.32\textwidth}
        \centering
        \includegraphics[width=1\linewidth]{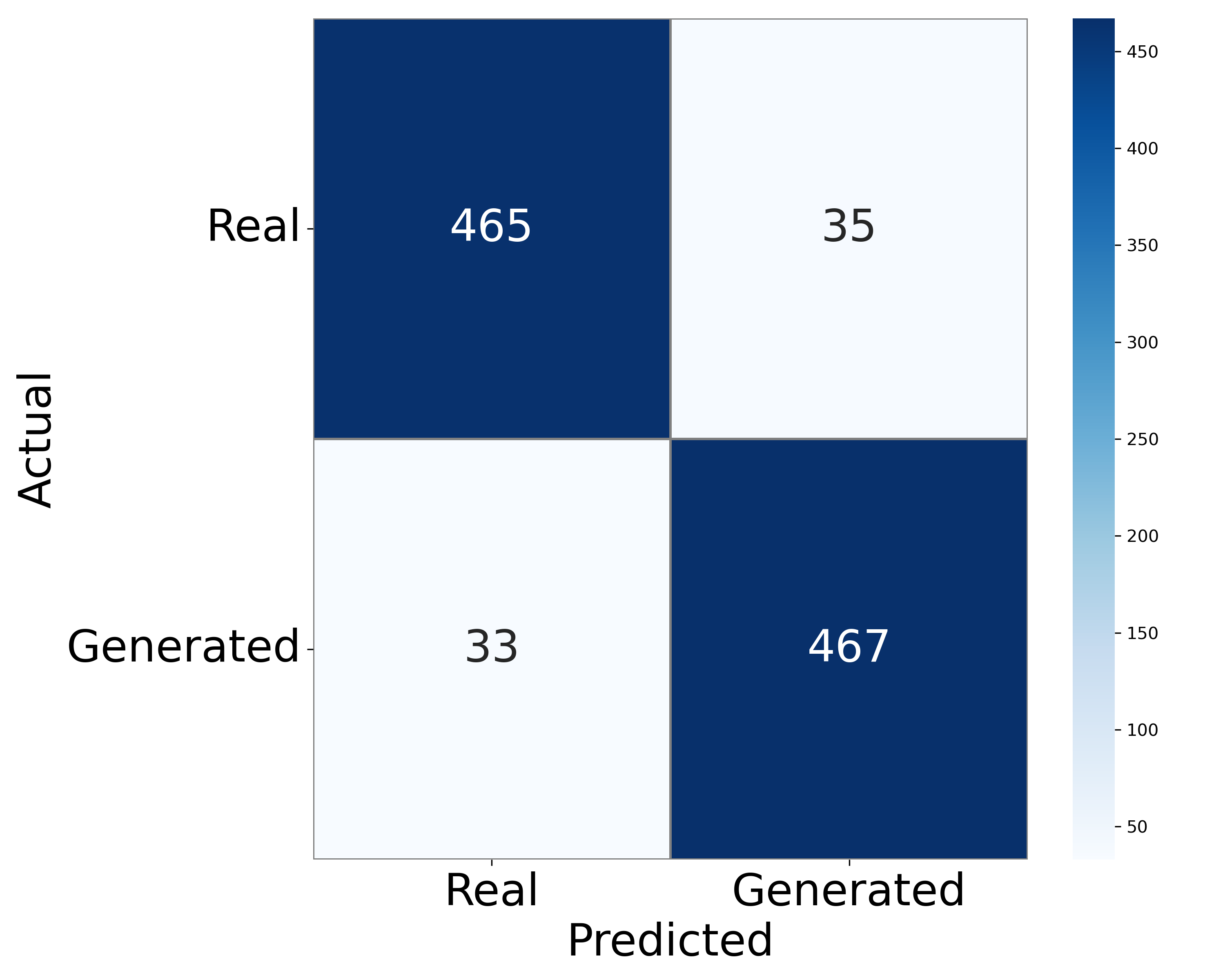}
        \caption{}
    \end{subfigure}%
    \begin{subfigure}[b]{0.32\textwidth}
        \centering
        \includegraphics[width=1\linewidth]{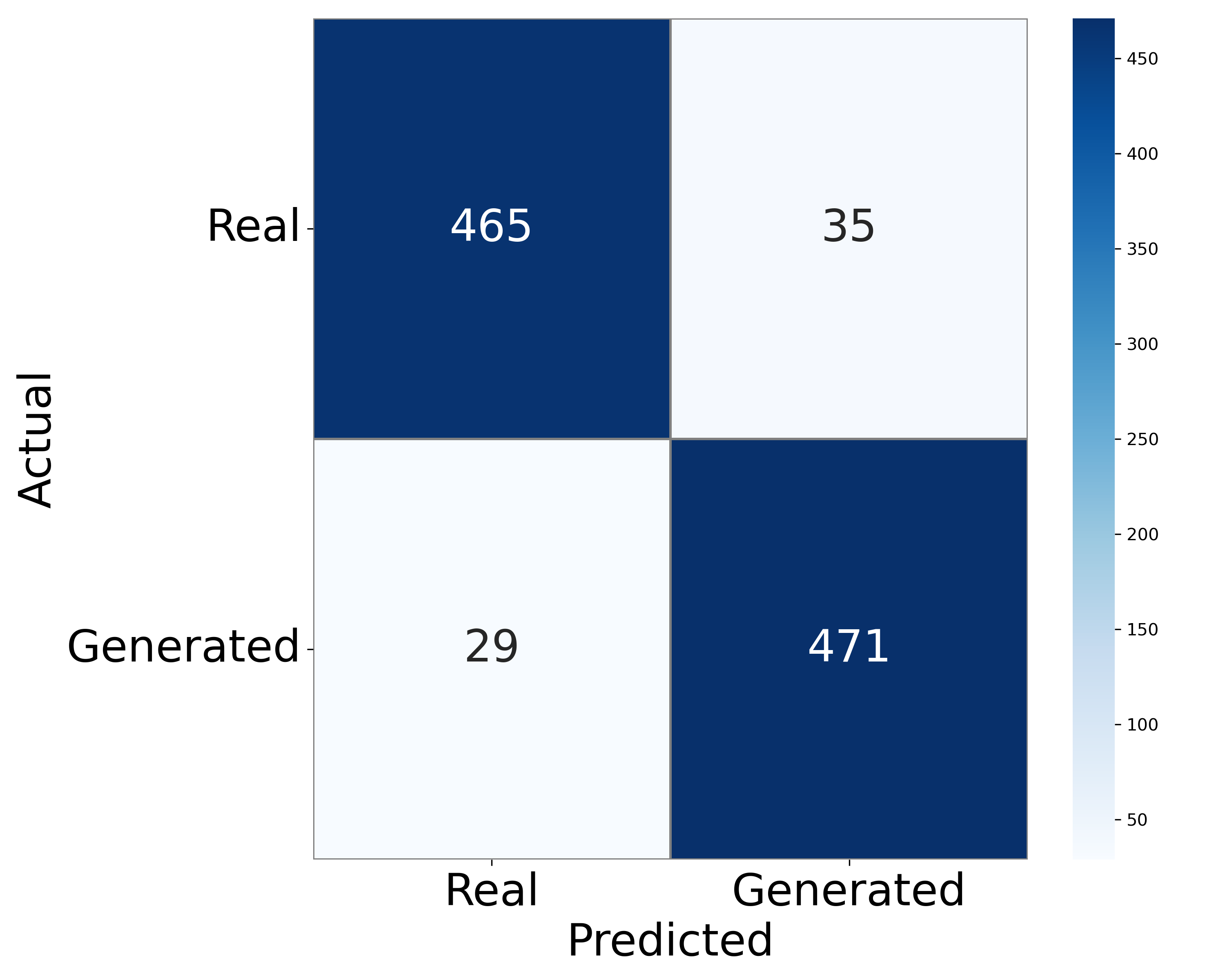}
        \caption{}
    \end{subfigure}%
    \begin{subfigure}[b]{0.32\textwidth}
        \centering
        \includegraphics[width=1\linewidth]{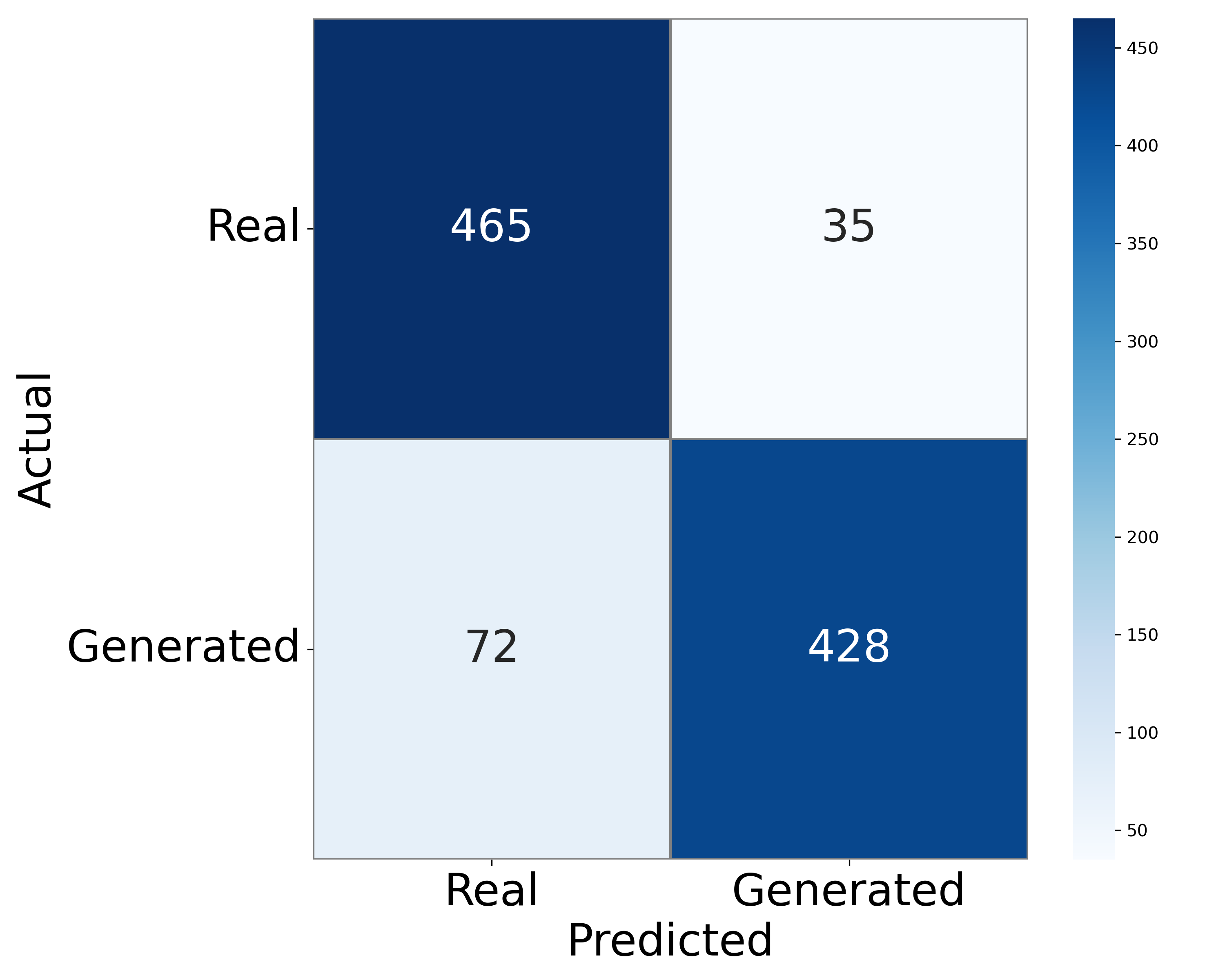}
        \caption{}
    \end{subfigure}
    \caption{Comparison of confusion matrices for the proposed hybrid classification approach and the baseline approach across three datasets. Subfigures (a), (b), and (c) depict the performance of the proposed approach on the Real vs. DALL-E 3, Real vs. MidJourney, and Real vs. Adobe Firefly datasets, respectively. Subfigures (d), (e), and (f) show the performance of the baseline MLP approach on the same datasets. These matrices illustrate the effectiveness of each approach in distinguishing real images from AI-generated images, with each matrix providing insights into the true positive, false positive, true negative, and false negative rates achieved.}
    \label{fig:CM}
\end{figure*}

In this section, we evaluate the performance of the Proposed Hybrid KAN-MLP and the Baseline MLP classifiers across three challenging Out of Distribution (OOD) datasets—Real vs. DALL-E 3, Real vs. Midjourney 5, and Real vs. Adobe Firefly. Each dataset consists of 1000 images, split evenly between real and AI-generated images, providing a comprehensive basis for assessing the robustness of our classifiers against OOD test data.  Precision, recall, and F1-scores for each condition are tabulated in Table \ref{tab:combined}, demonstrating the quantitative outcomes of our experiments. These metrics indicate the high level of accuracy achieved by the Proposed Hybrid KAN-MLP classifier, particularly in terms of F1-score, which is crucial for balancing the trade-off between precision and recall. Notably, the Proposed classifier consistently outperforms the Baseline across all datasets, affirming the effectiveness of the KANLinear module in enhancing feature discriminability and classification reliability.  The results from these evaluations are visually represented through confusion matrices and summarized in terms of precision, recall, and F1-score, which are key metrics for classifier performance assessment. The confusion matrices for each classifier and dataset combination are illustrated in Figure \ref{fig:CM}, which depicts the classifiers' ability to distinguish between real and generated images under varying conditions. 
\begin{table}[h]
\centering
\caption{Classification Results for Real vs. Generated Images Using Proposed and Baseline Approaches. The table presents a comparative analysis of the performance metrics — precision, recall, and F1-score — for both the proposed Hybrid KAN-MLP and the baseline MLP model across three OOD test datasets: DALL-E 3, MidJourney 5, and Adobe Firefly. }
\label{tab:combined}
\begin{tabular}{@{}>{\centering\arraybackslash}p{1cm}c>{\centering\arraybackslash}p{0.8cm}>{\centering\arraybackslash}p{0.8cm}>{\centering\arraybackslash}p{0.8cm}>{\centering\arraybackslash}p{0.8cm}>{\centering\arraybackslash}p{0.8cm}@{}} 
\toprule
\textbf{OOD Test Dataset} & \textbf{Approach} & \textbf{Class} & \textbf{Precision} & \textbf{Recall} & \textbf{F1-Score} & \textbf{Support} \\
\midrule
\multirow{4}{*}{Dall 3}& \multirow{2}{*}{Proposed} & Real & 0.97 & 0.92 & \textbf{0.94} & 500 \\
                                &                           & Gen& 0.92 & 0.97 & \textbf{0.95} & 500 \\
\cmidrule{2-7}
                                & \multirow{2}{*}{Baseline} & Real & 0.93& 0.93& 0.93 & 500 \\
                                &                           & Gen& 0.93 & 0.93& 0.93 & 500 \\
\midrule
\multirow{4}{*}{MJ 5}& \multirow{2}{*}{Proposed} & Real & 0.97 & 0.92& \textbf{0.94}& 500 \\
                                      &                           & Gen& 0.92& 0.97 & \textbf{0.95}& 500 \\
\cmidrule{2-7}
                                      & \multirow{2}{*}{Baseline} & Real & 0.94 & 0.92 & 0.93 & 500 \\
                                      &                           & Gen& 0.93 & 0.94 & 0.93 & 500 \\
\midrule
\multirow{4}{*}{Firefly}& \multirow{2}{*}{Proposed} & Real & 0.91 & 0.92 & \textbf{0.91} & 500 \\
                                        &                           & Gen& 0.91 & 0.90 & \textbf{0.91} & 500 \\
\cmidrule{2-7}
                                        & \multirow{2}{*}{Baseline} & Real & 0.87 & 0.92& 0.89 & 500 \\
                                        &                           & Gen& 0.92 & 0.86 & 0.89 & 500 \\
\bottomrule
\end{tabular}
\end{table}

\begin{figure*}
    \centering
    \begin{subfigure}[b]{0.32\textwidth}
        \centering
        \includegraphics[width=1\linewidth]{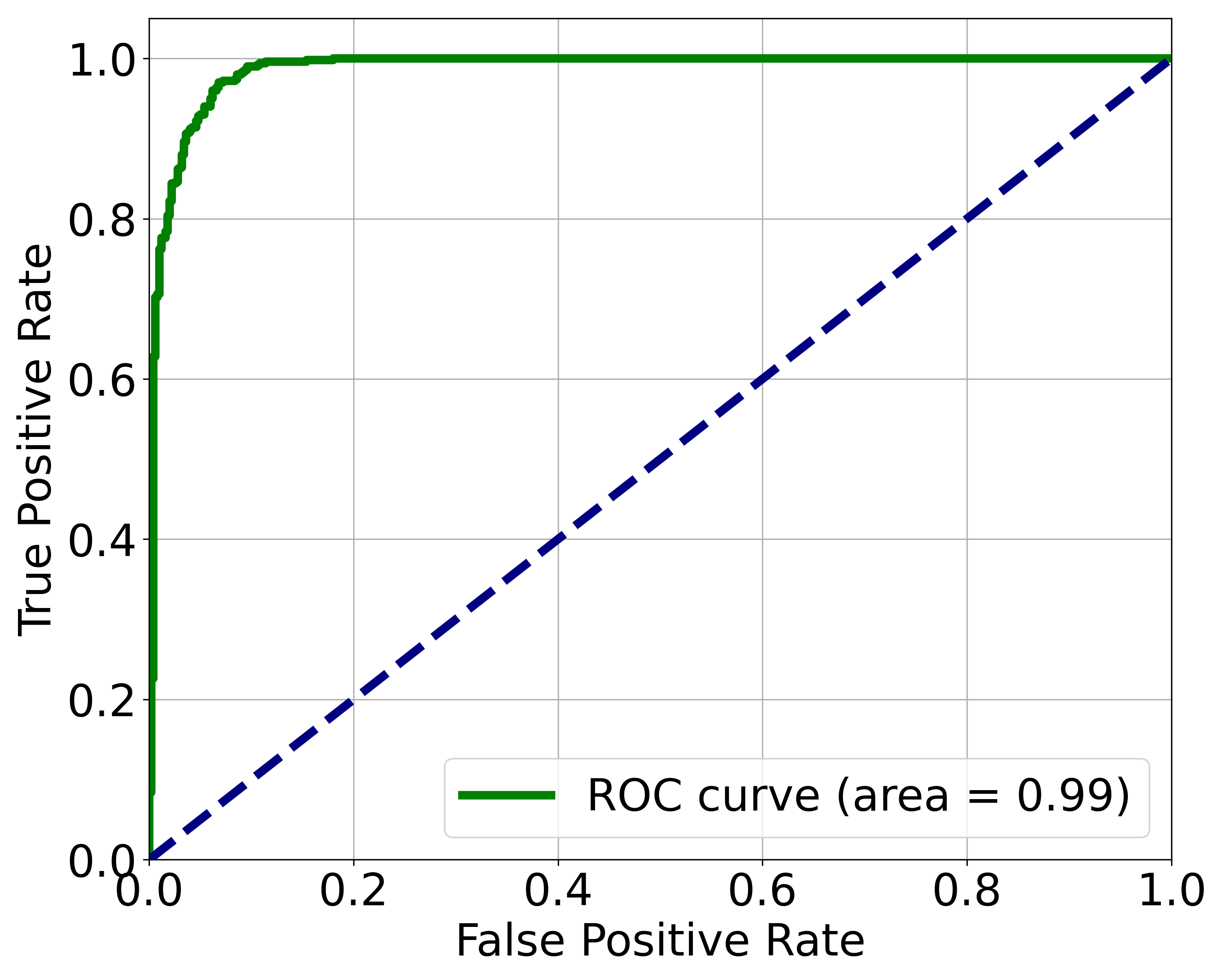}
        \caption{}
    \end{subfigure}%
    \begin{subfigure}[b]{0.32\textwidth}
        \centering
        \includegraphics[width=1\linewidth]{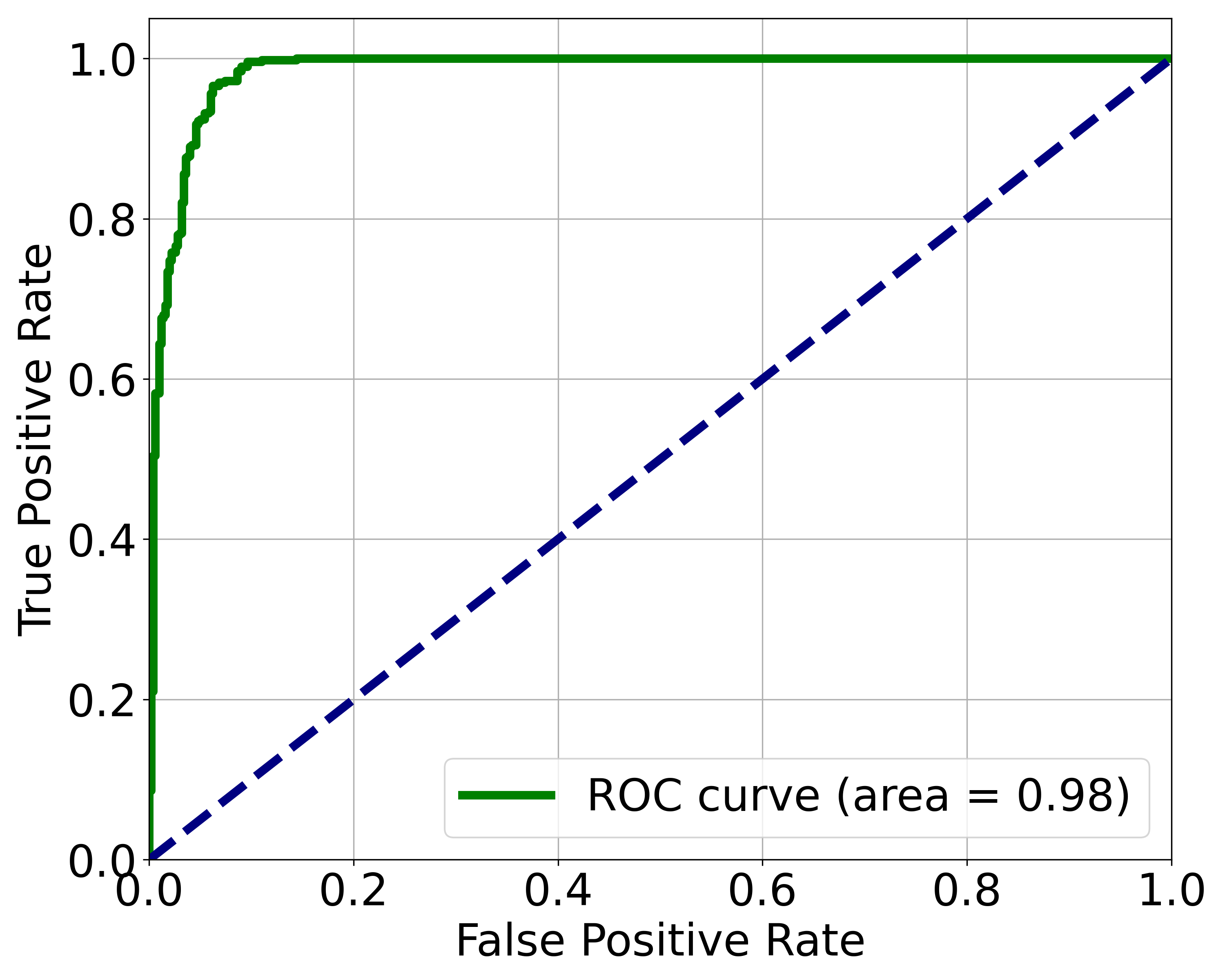}
        \caption{}
    \end{subfigure}%
    \begin{subfigure}[b]{0.32\textwidth}
        \centering
        \includegraphics[width=1\linewidth]{figures/AUC_ROC_Proposed_mid.png}
        \caption{}
    \end{subfigure}
    \caption{ROC curves illustrating the performance of the proposed hybrid classification approach on three datasets: (a) Real vs. DALL-E 3, (b) Real vs. MidJourney, and (c) Real vs. MidJourney. These curves demonstrate the classifier's discriminative ability by showing the trade-off between true positive rate (sensitivity) and false positive rate (1-specificity) across different thresholds, highlighting the model's effectiveness in accurately distinguishing between real and AI-generated images. The area under the curve (AUC) provides a quantitative measure of the overall performance of the classifier across these varied testing scenarios.}
    \label{fig:auc}
\end{figure*}

Additionally, the AUC ROC curves presented in figure \ref{fig:auc}  further substantiate the robust discriminatory power of our model. The curves indicate a significant separation between the true positive rate and false positive rate, underscoring the model’s efficiency in classification tasks under challenging conditions imposed by high-quality AI-generated images.Overall, the results corroborate the Hybrid KAN-MLP's potential as a formidable tool in fields requiring precise image validation, such as digital forensics and media integrity. The comparative analysis with the baseline also emphasizes the advancements incorporated through the KAN architecture, which significantly enhance the classifier's performance and adaptability. Future work will focus on scaling these approaches to accommodate larger datasets and integrating real-time processing capabilities, ensuring that the model remains effective against evolving generative image technologies.  

\section{Conclusion}
This study has successfully demonstrated the efficacy of the Hybrid KAN MLP model in distinguishing between real and AI-generated images, a crucial capability in the era of advanced digital image synthesis technologies. The robust performance of our classifier across diverse datasets underscores its potential utility in critical applications such as digital forensics and media integrity verification. The significant capability of our model to accurately classify images supports its potential integration into systems aimed at combating digital misinformation and ensuring the credibility of visual content in digital media. This is particularly relevant in today’s context, where the authenticity of digital content is frequently questioned. 

Despite its strong performance across various datasets featuring the latest AI-generated models like DALL-E 3, MidJourney 6, and Stable Diffusion 3 Ultra, our model faces a notable challenge: the high cost of using advanced AI services limits our ability to gather large datasets.  This limitation affects our ability to capture a wide range of behaviors from different AI models, making it important to manage dataset size carefully without compromising the model’s effectiveness. This is crucial as we plan to scale up the assessment of our classifier.  The model’s potential in areas like digital forensics, media verification, and content moderation remains significant. These fields rely on the accuracy of image analysis to maintain the trustworthiness of digital content. Moving forward, we aim to improve how our model handles large datasets cost-effectively and efficiently, preparing it to adapt quickly to new AI technologies while staying reliable and accurate.


\begin{thebibliography}{00}
\bibitem{pmlr-v139-ramesh21a}
A. Ramesh, M. Pavlov, G. Goh, S. Gray, C. Voss, A. Radford, M. Chen, and I. Sutskever,
``Zero-Shot Text-to-Image Generation,''
in \emph{Proceedings of the 38th International Conference on Machine Learning},
vol. 139, M. Meila and T. Zhang, Eds., PMLR, July 18--24, 2021, pp. 8821--8831.

\bibitem{Naitali2023}
A. Naitali, M. Ridouani, F. Salahdine, and N. Kaabouch, 
``Deepfake Attacks: Generation, Detection, Datasets, Challenges, and Research Directions,'' 
\emph{Computers}, vol. 12, no. 10, 2023, p. 216.
\url{https://www.mdpi.com/2073-431X/12/10/216}

\bibitem{lu2024seeing}
Z. Lu, D. Huang, L. Bai, J. Qu, C. Wu, X. Liu, and W. Ouyang,
``Seeing is not always believing: benchmarking human and model perception of AI-generated images,''
\emph{Advances in Neural Information Processing Systems},
vol. 36, 2024.

\bibitem{epstein_online_2023} 
D. C. Epstein, I. Jain, O. Wang, and R. Zhang, 
``Online Detection of AI-Generated Images,'' 
in \emph{ICCVW}, 2023.

\bibitem{chen_single_2023} 
J. Chen, J. Yao, and L. Niu, 
``A Single Simple Patch is All You Need for AI-Generated Image Detection,'' 
in \emph{ICCV}, 2023.

\bibitem{NEURIPS2021_49ad23d1}
P. Dhariwal and A. Nichol,
``Diffusion Models Beat GANs on Image Synthesis,''
in \emph{Advances in Neural Information Processing Systems}, vol. 34, 
M. Ranzato, A. Beygelzimer, Y. Dauphin, P.S. Liang, J. Wortman Vaughan, Eds.,
Curran Associates, Inc., 2021, pp. 8780--8794.
\url{https://proceedings.neurips.cc/paper_files/paper/2021/file/49ad23d1ec9fa4bd8d77d02681df5cfa-Paper.pdf}

\bibitem{wang_dire_2023} 
Z. Wang, J. Bao, W. Zhou, W. Wang, H. Hu, H. Chen, and H. Li, 
``DIRE for Diffusion-Generated Image Detection,'' 
in \emph{ICCV}, 2023.

\bibitem{Tsai2011}
M.-H. Tsai, J. Wang, T. Zhang, Y. Gong, and T. S. Huang,
``Learning Semantic Embedding at a Large Scale,''
in \emph{Proceedings of the 18th IEEE International Conference on Image Processing (ICIP)},
2011, pp. 2497--2500, doi: 10.1109/ICIP.2011.6116168.

\bibitem{yan_sanity_2024} 
S. Yan, O. Li, J. Cai, Y. Hao, X. Jiang, Y. Hu, and W. Xie, 
``A Sanity Check for AI-generated Image Detection,'' 
in \emph{IEEE Access}, 2024.

\bibitem{martin-rodriguez_detection_2023} 
F. Martin-Rodriguez, R. Garcia-Mojon, and M. Fernandez-Barciela, 
``Detection of AI-Created Images Using Pixel-Wise Feature Extraction and Convolutional Neural Networks,'' 
\emph{Sensors}, vol. 23, no. 23, 2023.

\bibitem{chai_what_2020} 
L. Chai, D. Bau, S.-N. Lim, and P. Isola, 
``What Makes Fake Images Detectable? Understanding Properties that Generalize,'' 
\emph{arXiv preprint arXiv:2007.01782}, 2020.

\bibitem{park_performance_2024} 
D. Park, H. Na, and D. Choi, 
``Performance Comparison and Visualization of AI-Generated-Image Detection Methods,'' 
in \emph{IEEE Access}, 2024.

\bibitem{DangNguyen2015RAISE}
D. T. Dang-Nguyen, C. Pasquini, V. Conotter, and G. Boato,
``RAISE: A Raw Images Dataset for Digital Image Forensics,''
in \emph{Proc. ACM Multimedia Systems},
Portland, Oregon, USA, March 2015.

\bibitem{esser2024scaling}
P. Esser, S. Kulal, A. Blattmann, R. Entezari, J. Müller, H. Saini, Y. Levi, D. Lorenz, A. Sauer, F. Boesel, et al.,
``Scaling Rectified Flow Transformers for High-Resolution Image Synthesis,''
in \emph{Proceedings of the Forty-first International Conference on Machine Learning}, 2024.

\bibitem{betker2023improving}
J. Betker, G. Goh, L. Jing, T. Brooks, J. Wang, L. Li, L. Ouyang, J. Zhuang, J. Lee, Y. Guo, et al.,
``Improving Image Generation with Better Captions,''
\emph{Computer Science},
vol. 2, no. 3, 2023.
\url{https://cdn.openai.com/papers/dall-e-3.pdf}

\bibitem{radford2021learning}
A. Radford, J. W. Kim, C. Hallacy, A. Ramesh, G. Goh, S. Agarwal, G. Sastry, A. Askell, P. Mishkin, J. Clark, et al.,
``Learning Transferable Visual Models from Natural Language Supervision,''
in \emph{Proceedings of the International Conference on Machine Learning},
pages 8748--8763, PMLR, 2021.

\bibitem{liu2024kan}
Z. Liu, Y. Wang, S. Vaidya, F. Ruehle, J. Halverson, M. Solja{\v{c}}i{\'c}, T. Y. Hou, and M. Tegmark, 
``KAN: Kolmogorov-Arnold Networks,'' 
arXiv preprint arXiv:2404.19756 (2024).

\end{thebibliography}
\end{document}